\documentclass[10pt,twocolumn,letterpaper]{article}

\usepackage{wacv}
\makeatletter
\@namedef{ver@everyshi.sty}{}
\makeatother
\usepackage{times}
\usepackage{epsfig}
\usepackage{graphicx}
\usepackage{amsmath}
\usepackage{amssymb}
\usepackage{booktabs}
\usepackage{tikz}
\usepackage{comment}
\usepackage{color}
\usepackage{makecell}
\usepackage{multirow}
\usepackage{xcolor}
\usepackage{tabularx,verbatim}
\usepackage{xspace}
\usepackage{caption}
\usepackage{tabu}
\usepackage{pifont}
\usepackage{enumitem}
\usepackage[accsupp]{axessibility} 
\newcommand{\xmark}{\ding{55}}

\usepackage[ruled,vlined]{algorithm2e}
\SetArgSty{textnormal}

\let\oldnl\nl
\newcommand{\nonl}{\renewcommand{\nl}{\let\nl\oldnl}}
\usepackage{bbm}
\usepackage{setspace}
\usepackage{comment}
\usepackage{wrapfig}
\usepackage{color, colortbl}

\definecolor{Gray}{gray}{0.92}
\definecolor{LightCyan}{rgb}{0.88,1,1}
\newcolumntype{L}{>{\arraybackslash}m{2cm}}

\newcommand{\domours}{D-CGCT\xspace}

\DeclareMathOperator*{\argmax}{arg\,max}

\def\wacvPaperID{365} 

\wacvalgorithmstrack   

\wacvfinalcopy 

\ifwacvfinal
\usepackage[breaklinks=true,bookmarks=false]{hyperref}
\else
\usepackage[pagebackref=true,breaklinks=true,colorlinks,bookmarks=false]{hyperref}
\fi

\begin{document}


\title{\textit{CoNMix} for Source-free Single and Multi-target Domain Adaptation}

\author{Vikash Kumar\thanks{Equal Contribution}~~~~
Rohit Lal$^\ast$~~~~
Himanshu Patil~~~~
Anirban Chakraborty\\
Indian Institute of Science, Bengaluru, India\\
{\tt\small \{vikashks,anirban\}@iisc.ac.in, \{take2rohit, hipatil1998\}@gmail.com}
}

\maketitle
\thispagestyle{empty}

\begin{abstract}
   This work introduces the novel task of \textbf{S}ource-\textbf{f}ree \textbf{M}ulti-\textbf{t}arget \textbf{D}omain \textbf{A}daptation and proposes adaptation framework comprising of \textbf{Co}nsistency with \textbf{N}uclear-Norm Maximization and \textbf{Mix}Up knowledge distillation (\textit{CoNMix}) as a solution to this problem. 
   The main motive of this work is to solve for Single and Multi target Domain Adaptation (SMTDA) for the source-free paradigm, which enforces a constraint where the labeled source data is not available during target adaptation due to various privacy-related restrictions on data sharing. The source-free approach leverages target pseudo labels, which can be noisy, to improve the target adaptation. We introduce consistency between label preserving augmentations and utilize pseudo label refinement methods to reduce noisy pseudo labels. Further, we propose novel MixUp Knowledge Distillation (MKD) for better generalization on multiple target domains using various source-free STDA models.
   We also show that the Vision Transformer (VT) backbone gives better feature representation with improved domain transferability and class discriminability. Our proposed framework achieves the state-of-the-art (SOTA) results in various paradigms of source-free STDA and MTDA settings on popular domain adaptation datasets like Office-Home, Office-Caltech, and DomainNet. Project Page: https://sites.google.com/view/conmix-vcl
\end{abstract}

\section{Introduction}

The advent of Deep Learning has brought significant development in tasks like image classification, object detection, semantic segmentation, etc. However, the performance of the state-of-the-art methods trained with millions of labeled images suffers significantly in the environment where there is a mismatch between training and test distribution ~\cite{imagenetv2,taori2020when}, motivating researchers to design learning algorithms that are robust to shifts in data distribution. One such popular research direction is Unsupervised Domain Adaptation (UDA) for a labeled source domain to a unlabeled target domain adaptation. UDA with only one source and one target domain is termed Single Target Domain Adaptation (STDA) \cite{wilson2020survey}. Multi-target Domain Adaptation (MTDA) consists of multiple unlabeled target domains against a single labeled source.
STDA can be thought of as a special case of MTDA and is critical in solving a practical task such as adaptation from Synthetic data distribution to Real-World data distribution. 
In contrast, the MTDA framework is essential when we have multiple target domains with varying domain-shift.

\begin{figure}[!htp]
\includegraphics[width=\linewidth]{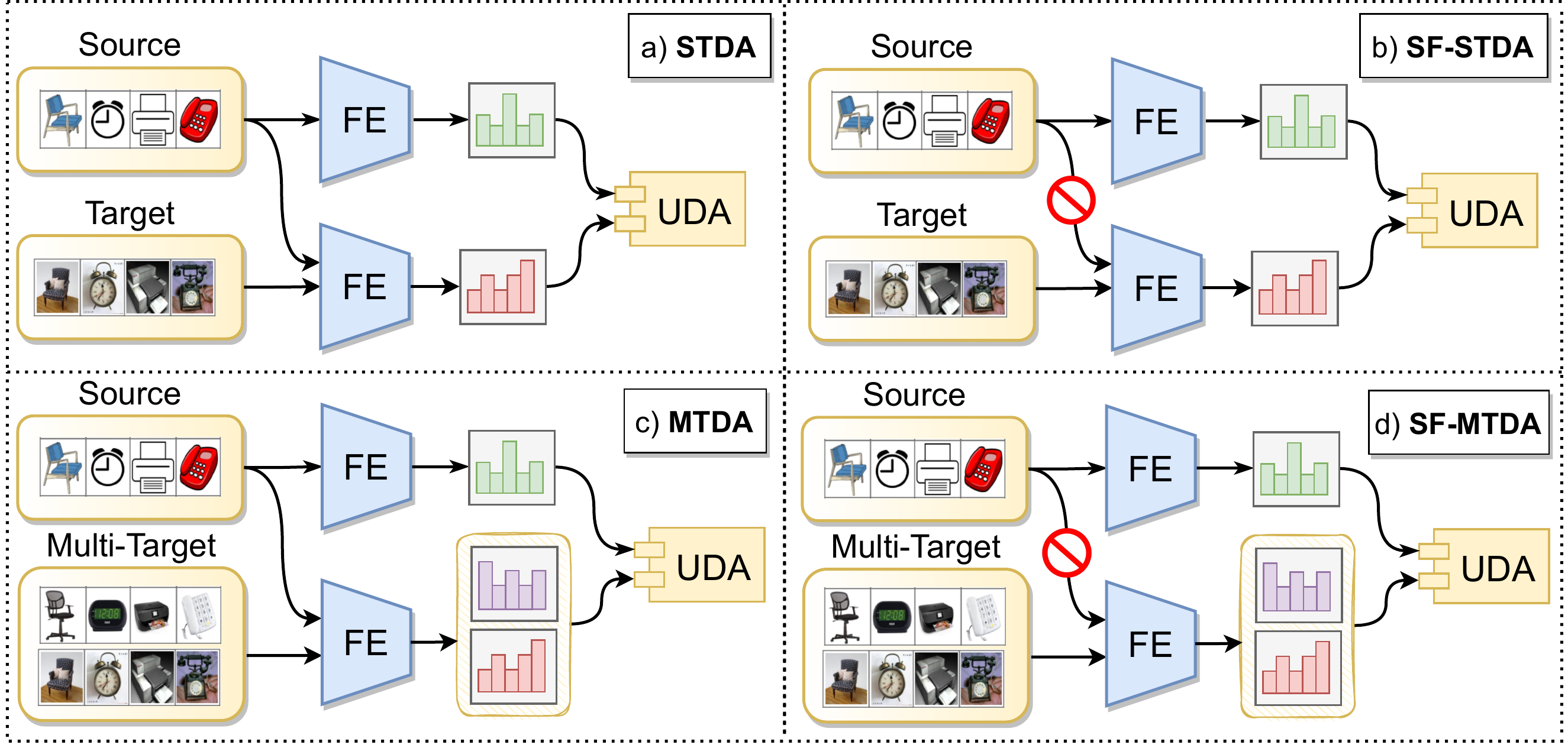}
\caption{\small\textbf{1.(a)} represents vanilla STDA approach where along with unlabeled target data, labeled source data and source trained model are available during adaptation stage.~\textbf{1.(b)} represents the adaptation strategy when only source trained model is available, but labeled source data is not available during adaptation stage. \textbf{1.(c)} is an extension of Fig1.(a), which shows an approach for UDA for the single source (always available) to multi-target domain adaptation (MTDA). \textbf{1.(d) (Ours)} is an extension to MTDA but without the access of labeled source data. Fig 1.(a), 1.(b), 1.(c) are already widely studied but 1.(d) remains unexplored.}
\label{fig:compare_methods}
\end{figure}

\begin{figure*}[!t]
\centering
\includegraphics[width=\linewidth]{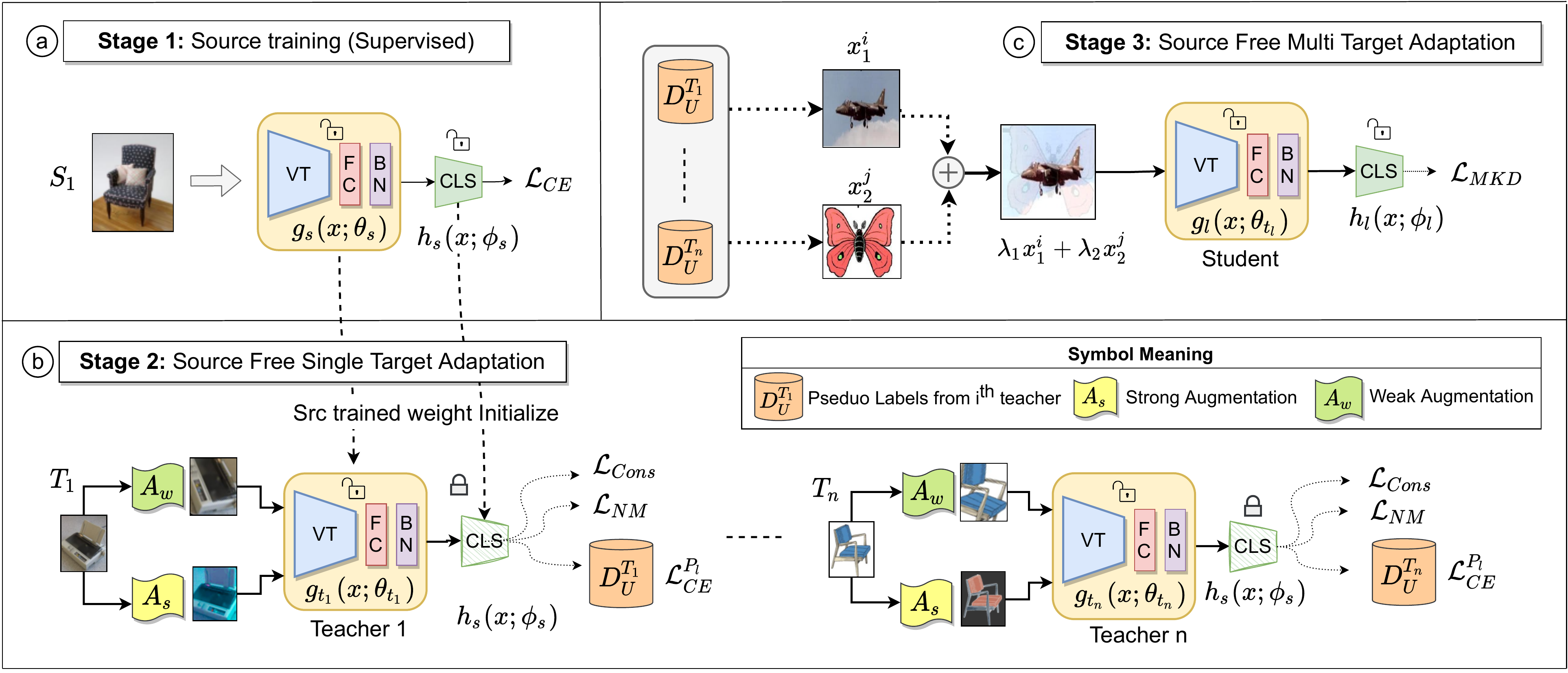}
\caption{\small \textbf{Architecture of \textit{CoNMix}}.
In Stage-1, we prepare a source trained model and get rid of source data. In Stage-2, we use previously source trained model and adapt to multiple target domains without having access to source. We store pseudo labels for all the target domains with the help of single domain adapted models (teachers) obtained in Stage-2. Finally, we perform knowledge distillation to a common student in Stage-3.}
\label{fig:sfmtda_architecture}
\end{figure*}

Most of the existing Domain Adaptation (DA) methods assume availability of labeled source domain samples during adaptation which may not be possible for several use cases that mandate data privacy, such as biometrics, healthcare etc. Also, there can be situations where storing a large dataset is not feasible, for instance, training and deploying domain adaptation applications on an embedded system or on edge devices with limited memory. However, we can store source trained models because they are relatively smaller in size compared to the dataset they are trained with. Therefore, traditional domain adaptation methods are not suitable in these situations and hence, we should look for \textit{source-free} methods. Source-free UDA methods \cite{kundu2020universal,liang2020we,liu2021source,kurmi2021domain,yang2021transformer} aims to solve the adaptation problem without access to labeled or unlabeled source data during adaptation to a target domain. We show the high-level distinction of our proposed approach with respect to existing unsupervised domain adaptation tasks  in Fig. \ref{fig:compare_methods}.

Source-free STDA works well for specified target domain; however, it may fail for many practical use-cases where the test data can come from multiple target domains or some unseen/open domains which were not present during adaptation. One trivial extension would be to train specific source-free STDA models for each target domain. In that case, we need to have the domain information for selecting the appropriate model during inference. Still with this constraint, it may not generalize well for open domains. One such practical application is scene classification in autonomous driving built by taking data from sunny weather that should generalise well across different weather conditions such as winter, rainy, foggy, etc where target labels are not available. To alleviate the above problems, we propose a novel source-free MTDA setting. In SF-MTDA setting, there would be a single trained model that can generalize well across different domains. This would reduce storage costs and have a shorter inference time than having a specific model for each target domain.

SF-MTDA poses the additional challenge of bridging the gap between multiple target domains by learning common representation. Existing state-of-the-art MTDA methods \cite{roy2021curriculum,nguyen2021unsupervised} use a complex adversarial training strategy that needs access to both source and target datasets for learning domain invariant representation making them unsuitable for source-free tasks. We propose to leverage pseudo-label refinement \cite{zhang2021refining} along with novel consistency constraint for mitigating the uncertainty associated with the pseudo-labels. To address SF-MTDA setting, we propose MixUp based Knowledge Distillation (MKD) to distill knowledge from multiple expert teachers (STDA models) to a single student. The overall novel framework for solving source-free domain adaptation tasks is dubbed as \textit{CoNMix}. Further, we also investigate the role of different backbones on source-free adaptation tasks.

Almost all existing UDA methods use CNN based feature extractors ~\cite{gholami2020unsupervised,chen2019blending,nguyen2021unsupervised,liu2020open,roy2021curriculum} whose design includes strong human inductive bias such as local connectivity and pooling.
Unlike CNN, VT has a global receptive field at every stage. Therefore, the learned representations are more meaningful for the downstream tasks. Self-attention in Vision Transformer is designed to assign more importance to salient objects of interest and lesser importance to less relevant information such as the background information. Therefore, it can mitigate the spurious correlation between prediction probability and domain dependent components such as lighting condition thereby making the feature representation more transferable, which is desirable in domain adaptation \cite{yang2021transformer}. In fact, in our experiments too, we observe that the feature representation of VT has better domain-transferability \textit{(easy to transfer across different domains)} and class-discriminability \textit{(ability to distinguish between classes)} compared to CNN based architecture (e.g.,ResNet).

\noindent In summary, our main contributions are as follows:
\begin{itemize}[noitemsep,topsep=0pt,leftmargin=*]
\item We propose a novel task of source-free multi-target domain adaptation and developed \textit{CoNMix}, a novel framework for solving source-free single and multi-target domain adaptation tasks. We also provide empirical insights, backed by quantitative and qualitative results to substantiate the use of VT backbone for SF-SMTDA.
\item We introduce a novel augmentation based consistency constraint and explore existing nuclear-norm maximization in our learning objective and pseudo label refinement strategy to mitigate the effect of noisy pseudo labels. Further, we judiciously combine these with Mixup knowledge distillation to propose the overall framework of CoNMix.
\item We are among the first to extensively study this important SF-MTDA problem. We have advanced the SOTA for SF-STDA and SF-MTDA settings on popular benchmark datasets. 
We also provide a new baseline on the large-scale DomainNet dataset for source-free single and multi-target domain adaptation.
\end{itemize}
The insights we draw from our analysis constitute important contribution of this paper. Compared to previous methods, CoNMix also has various appealing aspects- (a) Safe: CoNMix is developed to maintain complete data privacy, as it keeps the data safe and avoids any leakages (b) Flexible: A single algorithm can be extended for both source-free single and multi target domain adaptation (SF-SMTDA) tasks.

\section{Related Work}
\noindent\textbf{Single Target Domain Adaptation (STDA):} One of the popular methods for STDA is to try learning domain invariant features by minimising domain discrepancy \cite{yao2015semi,long2016unsupervised,sun2016return,long2017deep,kang2019contrastive}. Methods such as \cite{chen2020adversarial,Hoffman:Adda:CVPR17,long2018conditional,ganin2015unsupervised} leverage adversarial training for UDA. Generative modelling methods like \cite{bousmalis2017unsupervised,hong2018conditional} try to minimise the gap between source and target images by transforming one feature space to another. Though these methods have been proven to be very effective for STDA, their dependency on source data during adaptation makes it undesirable for source-free approaches.

\noindent\textbf{Source-free Domain Adaptation}: Recently source-free methods \cite{ahmed2021unsupervised,Li_2020_CVPR,liang2020we,kundu2020universal,yang2021transformer} are getting a lot of attention for UDA tasks. In this setting, we only have access to source trained model and unlabeled target data. Liang \etal \cite{liang2020we} uses information maximization and pseudo labeling to align the target domain to the source domain. 3C-GAN\cite{Li_2020_CVPR} improves the prediction through generated target-style data. Noisy pseudo label is one of the major problems in source-free adaptation tasks. 

\noindent\textbf{Multi Target Domain Adaptation}: For MTDA, we need to generalise for multiple unlabeled target data distribution with the help of single labeled source data distribution  \cite{gholami2020unsupervised,chen2019blending,liu2020open,peng2019domain}.Nguyen \etal~\cite{nguyen2021unsupervised} proposes training multiple adaptation networks and simultaneously distil knowledge from adapted models to small student network. The source-free MTDA is an important research direction, which has not been explored extensively yet to the best of our knowledge. Our proposed framework \textit{CoNMix} attempt to address SF-STDA and SF-MTDA problems.

\noindent\textbf{Vision Transformer (VT):} Transformer achieved a lot of success in natural language processing since it was first introduced by Vaswani \etal \cite{vaswani2017attention}. Dosovitskiy \etal \cite{dosovitskiy2020image} represented image patch with position encoding as a sequence dataset and reported improved performance on ImageNet. Touvron \etal \cite{touvron2021training} uses smaller dataset for training compared to \cite{dosovitskiy2020image} utilising distillation token for learning the inductive bias. Kurmi \etal \cite{kurmi2019attending} proposes to get the weighted feature representation by multiplying backbone output with the attention map generated through the Bayesian discriminator.  Due to the absence of source sample during adaptation, we can not use domain discriminator based architecture, therefore these methods can not be extended for source-free tasks. Yang \etal \cite{yang2021transformer} uses the bigger variant of Vision Transformer (ViT-B) \cite{dosovitskiy2020image} along with student-teacher architecture for solving SF-STDA problem. ViT-B is overparameterized with 86M parameters, whereas existing methods use ResNet50 which has only 24M parameters.  


\section{Problem Setting and Proposed Approach}

In this section we define the problem setting and our proposed approach towards solving this problem. We are trying to solve source-free single and multi-target domain adaptation, which involves solving adaptation task on single and multiple unlabeled target domain using source trained model without accessing the source dataset during adaptation. 

We introduce \textit{CoNMix} (Fig. \ref{fig:sfmtda_architecture}), a three-stage approach that utilizes Vision Transformer along with consistency constraint, nuclear norm maximization, pseudo label refinement and MixUp based knowledge distillation (MKD), designed at solving source-free single and multi-target domain adaptation problem. 

\noindent\textbf{Notation:} We denote $h$ as hypothesis or classifier. $\xi_S(h)$ and $\xi_T(h)$ are expected risk/error of hypothesis $h$ for source domain and target domain respectively. $\mathcal{L}_{CE},~\mathcal{L}_{NM},~\mathcal{L}_{Cons},~\mathcal{L}_{CE}^{P_l},~\mathcal{L}_{MKD}$ represents cross-entropy loss, Nuclear-norm Maximization loss, Consistency loss, Pseudo label Cross-Entropy loss, and MixUp Knowledge Distillation loss respectively. $\mathcal{A}_{W}$ and $\mathcal{A}_{S}$ are weak and strong augmentation applied to input sample. We use $\mathcal{X_S},~\mathcal{Y_S}, ~\mathcal{X_T}$ for representing source image, source label and target image respectively. $\mathcal{D_S}$ and $ ~\mathcal{D_T}$ are source and target distribution. ${d}_{\mathcal{H}\Delta\mathcal{H}}(\mathcal{D_S},\mathcal{D_T})$ is divergence between source and target domain distribution and  $\hat{d}_{\mathcal{H}\Delta\mathcal{H}}(\mathcal{X_S},\mathcal{X_T})$ is its empirical measure. 

\subsection{Backbone Selection}

\label{subsec:proposal_motivation}
In this section, we demonstrate that the attention based backbone \cite{dosovitskiy2020image} provide tighter upper bound on  $\xi_T(h)$ compared to popular ResNet \cite{he2016deep} backbone. We provide empirical insight to show that $\xi_{T}^{VT}(h) < \xi_{T}^{RN50}(h)$, making Vision Transformer (VT) a more suitable candidate for solving domain adaptation tasks. Additional information and comparison pertaining to the choice of backbone in our proposed approach are provided in the following subsections.

\subsubsection{Comparison of Backbone}
\begin{figure}[!htpt]
\includegraphics[width=0.8\linewidth]{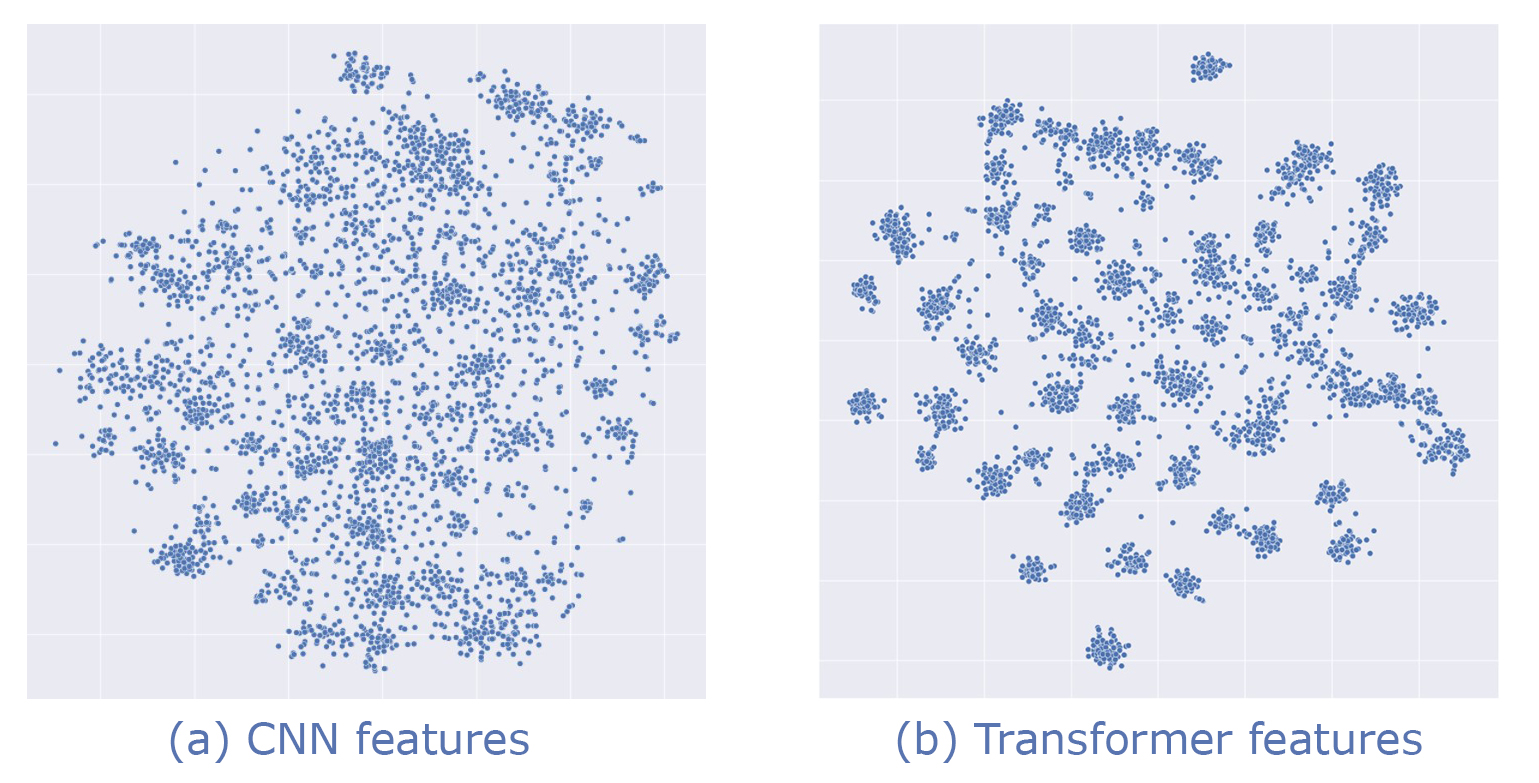}
        \centering
        \captionof{figure}{\small \textbf{t-SNE plots} (a) and (b)  are t-SNE plots using features of $Cl$ images obtained by passing through \textit{Rw} $\to$ \textit{Cl} adapted model for ResNet50 and VT respectively. }
        \label{fig:tsne}
\end{figure}

\begin{figure}[!h]
\includegraphics[width=0.95\linewidth]{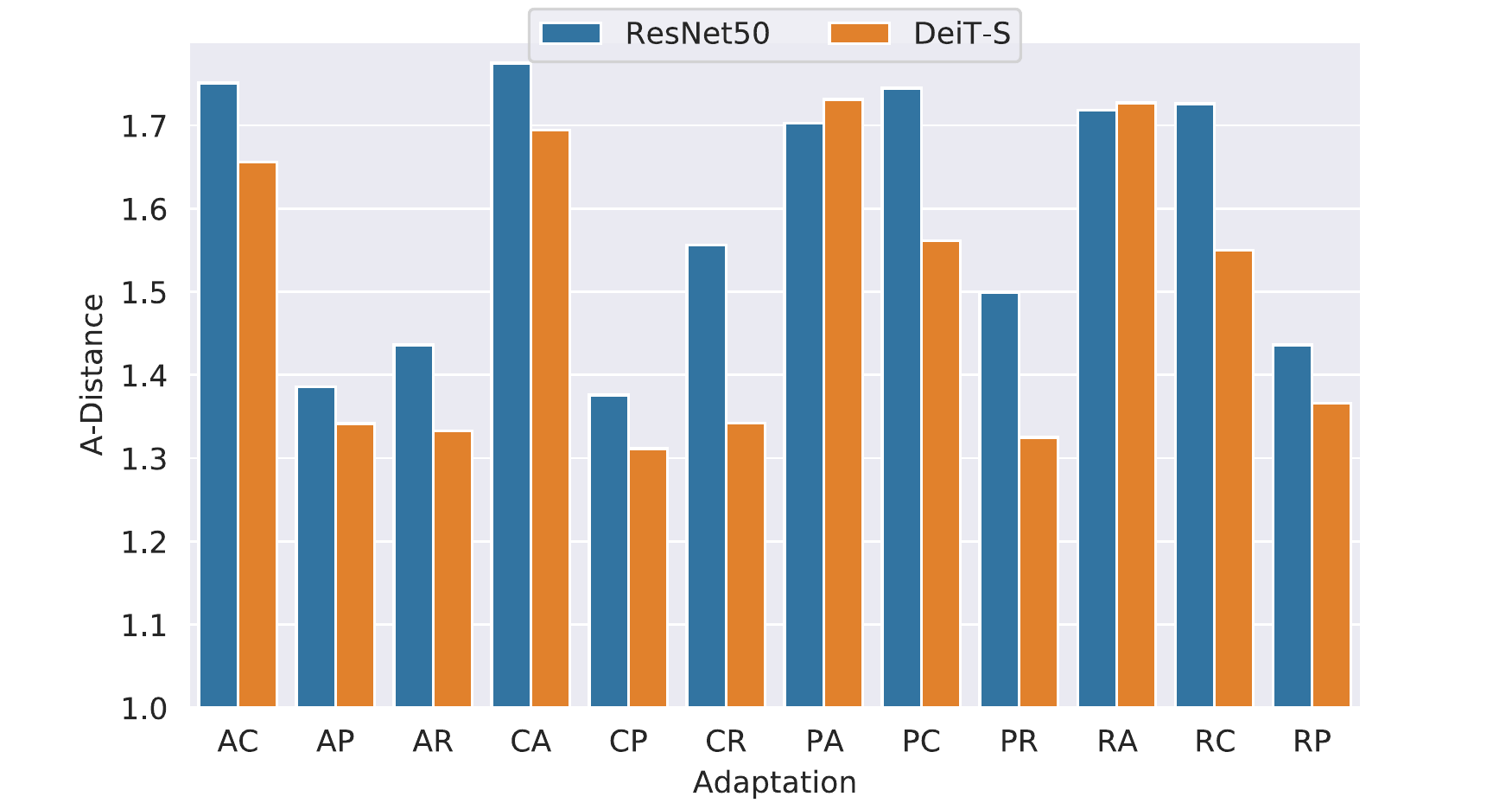}
        \captionof{figure}{\small Plot compares $\mathcal{A}$-distance of VT and RN50 backbone for various Office-home splits (e.g., AC represents adaptation from Art to Clipart). Small \textbf{$\mathcal{A}$-distance} results in better feature alignment. VT features consistently shows smaller $\mathcal{A}-distance$.}
        \label{fig:a_distance}
\end{figure}

A majority of current SOTA UDA techniques extract image features using a CNN based backbone, such as Resnet50. Given the recent success of VT \cite{dosovitskiy2020image,touvron2021training}, we attempt to analyse the feature representation of VT based backbone for domain adaptation. We aim to show difference in learned representations using RN50 and DeiT backbones. In our experiments, we find that \textit{VT features are more domain-transferable and class-discriminative compared to ResNet}. We corroborate above by explicitly measuring the $\mathcal{A}$-distance in Fig. \ref{fig:a_distance} which is a popular way to measure the feature alignment in adversarial learning \cite{ganin2015unsupervised}. We found that $\mathcal{A}$-distance of VT feature representation is smaller compared to CNN based representation. This difference in $\mathcal{A}$-distance shown in Fig. \ref{fig:a_distance} is significant and provides a direct evidence to why DeiT backbone leads to substatially better performance in UDA. Additionaly, we examined the t-SNE plot of the two representations and discovered that VT based representation are relatively better aligned (Fig. \ref{fig:tsne}). We believe that the properties of VT (DeiT-S) such as having the global receptive field at every stage and self-attention help them learn more class-discriminative and domain-transferable feature representation than CNN (ResNet).  

We also observe that \textit{our suggested loss functions are better suited for VT than ResNet}. In supp. material, we perform an experiment to analyse the effect of loss functions on two backbones. We discovered that VT backbone results in significantly increased performance compared to its CNN counterpart. Based on this analysis, we can conclude that VT serves as a better feature extractor alternative than CNN for domain adaptation tasks. We also compare the effect of various VT models and better ImageNet models like EfficientNetV2-B3 and EfficientNetV2-S in suppl. material to further validate our analysis. 

\subsection{Source-Free Domain Adaptation}
\label{subsec:sda}

The \textit{CoNMix} architecture attempt to solve two problems. Firstly, It can solve source-free Single-Target Domain Adaptation (SF-STDA) by utilizing Stage-1 and Stage-2 of the Architecture (Fig. \ref{fig:sfmtda_architecture}). Secondly, by introducing Stage-3, \textit{CoNMix} can solve source-free Multi-Target Domain Adaptation (SF-MTDA) effectively.
\subsubsection{Source Training}
\label{subsub:src_training}
We aim to learn a model $f_{s}(\theta_{s},\phi_{s}) : \mathcal{X_S} \rightarrow \mathcal{Y_S}$ where $\theta_{s},~\phi_{s}$ are parameter for backbone network and classifier network respectively. Differing from traditional approaches, additional fully connected layer and batch norm layer was used after the backbone for better alignment of the projected features. We use VT backbone due to it's discussed benefits over ResNet. We sample $(\mathcal{X_S}, ~\mathcal{Y_S}) \sim \mathcal{D_S}$ and train the complete network using cross-entropy $\mathcal{L}_{CE}$ loss as shown in Eq.\ref{eq:cross}.
\begin{equation}\small
		\mathcal{L}_{CE} = 
		-\mathbb{E}_{(x_s,y_s)} \sum\nolimits_{k=1}^{K} q_k \log \delta_k(f_s(x_s:\theta_{s},\phi_{s}))
	\label{eq:cross}
\end{equation}
where $\delta_k(a)=\frac{\exp(a_k)}{\sum_i \exp(a_i)}$ denotes the $k$-th element in the soft-max output of a $K$-dimensional vector $a$, and $q$ is the one-of-$K$ encoding of $y_s$ where $q_k$ is `1' for the correct class and `0` for the rest. In practice, we have used label smoothing \cite{muller2019does} in place of one-hot encoding to avoid the network being overconfident. It helps to improve model generalization ability. $q_k$ will be replaced with $q^{ls}_k=(1-\alpha)q_k + \alpha/K$ where $q^{ls}_k$ is the smoothed label and $\alpha$ is the smoothing parameter set to 0.1. Refer Supp. material for more discussion on effect of smooth labels on adaptation. Post source training, we freeze the classifier network. Please note that the \textit{labeled source sample will not be available for the next target adaptation stage} because we are trying to solve the source-free problem. 
\subsubsection{Source-free Single-target Domain Adaptation}
\label{subsub:sf_sta}
We have access to sampled unlabeled target data $(\mathcal{X_T}) \sim \mathcal{D_T}$ and source trained model $f_{s}(\theta_{s},\phi_{s})$. Stage-2 aims to train $T$ independent single source to single target domain adaptation networks while having no access to source data where $T$ is a total number of target domains. Following \cite{liang2020we}, we propose to freeze the classifier parameter $\phi_{s}$ and update the backbone parameter $\theta_{t}$ which was initialized from source backbone parameter $\theta_{s}$. We use $\mathcal{L}_{CE}^{P_l}$, $\mathcal{L}_{NM}$ and $\mathcal{L}_{Cons}$ for updating the backbone weights using back-propagation with SGD \cite{le2011optimization}. 

\noindent\textbf{Nuclear-norm Maximization (NM):} Many label insufficient situations such as Semi-supervised learning \cite{zhu2009introduction} or Unsupervised learning \cite{hastie2009unsupervised} suffers from higher data density near decision boundary, which results in poor class-discriminability. Directly minimizing the Shannon entropy  \cite{shannon2001mathematical} leads to uniformly smooth representation which improves discriminability by pushing samples to one of the class labels, however, it does not ensure diversity and may result in undesirable solution where all the minority class is pushed to the nearest majority class. Different variants such as Information maximization (IM) loss \cite{hjelm2018learning} address this issue with limited success. Nuclear-norm maximization (NM) \cite{cui2020towards} uses batch-statistics to achieve function-smoothing only in the required dimensions and to the required extent leading to superior representation therefore, it improves both class-discriminability as well as prediction diversity in a unified way making it desirable for SF-SMTDA tasks.

\noindent\textbf{Class-discriminability in NM:} We define $A \in \mathbb{R}^{B \times K}$ to be the classification-response matrix A, where $B$ is the batch size, and  $K$ is the number of classes. Frobenius norm  $\|A\|_F$ is defined in Eq.\ref{eq:frob_norm}.
\begin{equation} \label{eq:frob_norm}
    \begin{aligned}
	    ||A||_{F} = \sqrt{\sum_{i=1}^{B}\sum_{j=1}^{K}|A_{ij}|^{2}}
	\end{aligned}
\end{equation}
\noindent where $0 \leq A_{ij} \leq 1$ and $\sum_{j=1}^{K}(A_{ij})=1$. We can obtain the upper bound of $||A||_{F}$ as shown in Eq.\ref{eq:frob_norm_bound}
\begin{equation} \label{eq:frob_norm_bound}
    \begin{aligned}
	    ||A||_{F} \leq \sqrt{\sum_{i=1}^{B}(\sum_{j=1}^{K}A_{ij})(\sum_{j=1}^{K}A_{ij})}
	    =\sqrt{\sum_{i=1}^{B}(1.1)} = \sqrt{B}
	\end{aligned}
\end{equation}
\noindent Upper bound in $||A||_F$ corresponds to the one-hot prediction for each sample in a batch. Therefore, maximizing $||A||_F$ leads to improved class-discriminability. Cui \etal ~in \cite{cui2020gradually} proved that maximum value of $\|A\|_{F}$ comes where entropy achieves its minimum value.

\noindent\textbf{Prediction-diversity in NM:} If we define $\|A\|_{*}$ as nuclear norm and r as the rank of $A$ then Recht \etal \cite{recht2010guaranteed} provide relation between $||A||_F$ and $||A||_{*}$ , which we show in Eq. \ref{eq:nuc_from_rel}. We provide proof in supp. material.

\begin{equation} \label{eq:nuc_from_rel}
    \begin{aligned}
	    \|A\|_F \leq \|A\|_* \leq \sqrt{r}\|A\|_F 
	\end{aligned}
\end{equation}
\noindent If $K < B$, then the $r$ approximates the number of classes present in the batch by finding the linearly independent column vectors. Therefore, improving the rank of $A$ is desirable. Our objective is to maximize the rank ($r$) of $A$ which can be achieved by maximizing the nuclear-norm of it. Inequality shown in Eq.\ref{eq:nuc_from_rel} suggest that we can achieve the desired objective by maximizing the $||A||_F$ because it also provides the lower bound of $||A||_{*}$. Cui \etal shows that the approximation of NM using batch Frobenius norm improves the model performance and reduces the training time  \cite{cui2021fast}.  Hence, we define the Nuclear-Norm loss using its Frobenius approximation as shown in  Eq.\ref{eq:norm_loss}. 
\begin{equation} \label{eq:norm_loss}
    \begin{aligned}
	    \mathcal{L}_{NM} = -\|A\|_{F} = -\|(f_t(X_{T}^{B};\theta_{t},\phi_{s})\|_{F}
	\end{aligned}
\end{equation}
\noindent where $\|(f_t(X_{T}^{B};\theta_{t},\phi_{s})\|_{F}$ is Frobenius norm of classification-response matrix thereby minimizing $\mathcal{L}_{NM}$ improves class-discriminability as well as prediction-diversity.

\noindent\textbf{Initial Pseudo label (PL):} To improve model performance using self-training \cite{shin2020two}, we propose to use Pseudo label based cross-entropy loss ($\mathcal{L}_{CE}^{P_l}$ in Eq.\ref{eq:pseudo_ce}). Pseudo labels are inherently noisy, so directly computing target pseudo labels using source trained model is not desirable \cite{rizve2021defense}. We use it along with nuclear norm maximization, which acts as soft regularization for self-training. We follow an iterative strategy similar to Liang \etal \cite{liang2020we} to obtain pseudo labels. We get the initial class $c_k^{(init)}$ center using weighted k-means as shown in Eq.\ref{eq:k_means}
\begin{equation}
		c_k^{(init)} = \frac{\sum_{x_t\in \mathcal{X_T}}{\delta_k(f_t(x_t;\theta_{t},\phi_{s}))}\ g_t(x_t;\theta_{t})}{\sum_{x_t\in \mathcal{X_T}}{\delta_k(f_t(x_t;\theta_{t},\phi_{s}))}}
		\label{eq:k_means}
\end{equation}
We can find the pseudo label $\forall x_t \in \mathcal{X_T}$ based on their maximum cosine similarity with the initial class-center as shown in Eq.\ref{eq:cosine}.
\begin{equation}\label{eq:cosine}
		\hat{y}_{t}^{init} = \argmax_k\; \frac{\langle{g}_t(x_t;\theta_{t}), c_k^{(init)}\rangle}{\|{g}_t(x_t;\theta_{t})\|\|c_k^{(init)}\|}
\end{equation}
This allows us to assign each target sample to only one class. We can find the updated class center using the fraction of sample belonging to each class. We use Eq.\ref{eq:k_means} ~and Eq.\ref{eq:cosine} ~in an iterative manner to find the updated pseudo label.
\begin{figure}[!htp]
\includegraphics[scale=0.31]{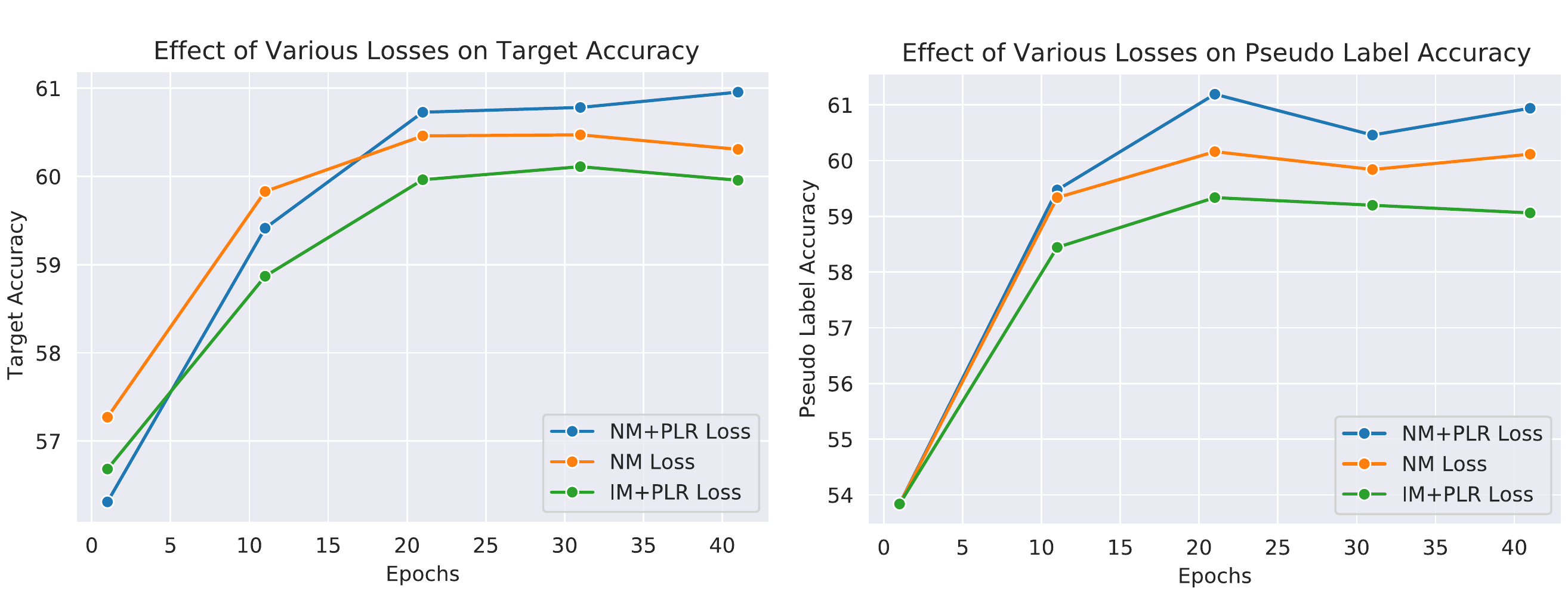}
\centering
\caption{\small Plot for Nuclear-Norm Maximization (NM) vs. Information Maximization (IM) for Ar$\to$Cl. \textit{Left} and \textit{Right} plot compares target accuracy and pseudo label accuracy for IM and NM with and without pseudo-label refinement (PLR) respectively. NM+PLR used in CoNMix provides the best performance.}
\label{fig:loss_ablation}
\end{figure}

\noindent\textbf{Pseudo Label Refinement (PLR):} In order to reduce the noise in pseudo label, we refine pseudo labels using the temporal ensemble and consensus based weighting scheme \cite{zhang2021refining}. Intuitively, If a pseudo label is consistent in two consecutive epochs then it should get more weight and vice-versa. Let $\tilde{y}_{z}^{n-1}$ and $\tilde{y}_{z}^{n}$ are pseudo label for $z^{th}$ sample in epoch $n$ and $n-1$ respectively. Let $W \in R^{K \times K}$ is cluster consensus matrix where, $K$ is the total number of classes. If total number of samples in $i^{th}$ class at $n^{th}$ epoch is denoted as $I^{n}(i)$ then $W(i, j)$ is shown in Eq.\ref{eq:label_refine} 
\begin{equation}\label{eq:label_refine}
    W(i,j) = \frac{|I^{n-1}(i) \cap I^{n}(j)|}{|I^{n-1}(i) \cup I^{n}(j)|} \in [0,1]
\end{equation}
Where $|.|$ is cardinality  of a set. Row normalized $W(i,j)$ captures the similarity between $i^{th}$ class and $j^{th}$ class in epoch $n-1$ and $n$. Ideally, off-diagonal entries of matrix $W$ should be close to zero. Finally, the updated pseudo label is shown in Eq.\ref{eq:final_label2}.
\begin{equation}\label{eq:final_label2}
{\hat{y}}_{z}^{n} = \alpha \tilde{y}_{z}^{n}  + (1 - \alpha)W^{T}\tilde{y}_{z}^{n-1}
\end{equation}
where $\alpha$ is a hyper-parameter.
We use refined pseudo label to calculate the 
$\mathcal{L}_{CE}^{P_l}$ loss as shown in Eq.\ref{eq:pseudo_ce} where $\hat{y}$ is refined pseudo label and $\mathbbm{1}_{[k=\hat{y}_t]}$ is indicator function. Fig. \ref{fig:loss_ablation} shows the effectiveness of PLR when used with NM and IM loss.
\small
\begin{equation}\label{eq:pseudo_ce}
	\begin{aligned}
	    \mathcal{L}_{CE}^{P_l} = -\mathbb{E}_{(x_t,\hat{y}_t)\in \mathcal{X_T} \times \hat{\mathcal{Y_T} }} & \sum\nolimits_{k=1}^{K}  \mathbbm{1}_{[k=\hat{y}_t]} \log \delta_k(f_t(x_t;\theta_{t},\phi_{s}))
	\end{aligned}
\end{equation}
\normalsize

\noindent\textbf{Consistency Loss:} 
For learning domain invariant representation, we propose weak and strong augmentation of the target image and seek consistent representation across the two label preserving augmentations as shown in Fig. \ref{fig:sfmtda_architecture}(b). Let, ${X}_{tw}^{B}, ~{X}_{ts}^{B}$ are weak and strong augmentation for target batch ${X}_{t}^{B}$ and ${Y}_{tw}^{B}, ~{Y}_{ts}^{B}$ are respective model softmax output i.e ${Y}_{tw}^{B} = \delta_k(h_{t}(g_{t}({X}_{tw}^{B}))$ and ${Y}_{ts}^{B} = \delta_k(h_{t}(g_{t}({X}_{ts}^{B}))$.
We define an expectation ratio as $\mathcal{E}_{ratio} = \mathbb{E}[{Y}_{tw}^{all}]/\mathbb{E}[{Y}_{tw}^{B}]$. 
${Y}_{tw}^{B}$ is then normalized as  $\Hat{{Y}}_{tw}^{B} = \delta_k({Y}_{tw}^{B}\mathcal{E}_{ratio})$ such that the row sum is 1.
$\Hat{{Y}}_{tw}^{B}$ acts as soft label ground truth for strong augmented output ${Y}_{ts}^{B}$ and we minimize soft label based cross-entropy loss as shown below.
\begin{equation} \label{consist_loss}
    \mathcal{L}_{cons} =
    -\mathbb{E}_{(y_{ts})\in {Y}_{ts}^{B}} \sum\nolimits_{k=1}^{K}\Hat{{y}}_{tw}^{k}  \log {y}_{ts}^{k}
\end{equation}  
\noindent $\mathcal{E}_{ratio}$ ensures that first order batch statistic matches with first order overall target data statistics. Overall loss for Stage-2 training is given by $\mathcal{L}_{total}$
\small
\begin{equation} \label{eq:stage_2_loss}
\mathcal{L}_{total} = \lambda_1 \mathcal{L}_{NM} + \lambda_2 \mathcal{{L}}_{CE}^{P_l} + \lambda_3 \mathcal{L}_{cons} 
\end{equation}
\normalsize
where $\lambda_1, \lambda_2, \lambda_3$ are weights associated w.r.t three losses $\mathcal{L}_{NM}, \mathcal{L}_{CE}^{P_l}$ and $\mathcal{L}_{cons}$ respectively. In Fig. \ref{fig:loss_ablation} we show that accuracy of pseudo label increases as training progresses which will result in improved adaptation. IM loss has been used in source-free domain adaptation task \cite{liang2020we,Spp} whereas NM loss is not explored for this task.

\begin{table*}[!htp]
\centering
\setlength{\tabcolsep}{3pt}
\resizebox{0.7\linewidth}{!}{%
\begin{tabular}{l|c|ccccccccccccc}
\toprule
Method & SF & Ar$\to$Cl & Ar$\to$Pr & Ar$\to$Rw & Cl$\to$Ar & Cl$\to$Pr & Cl$\to$Rw & Pr$\to$Ar & Pr$\to$Cl & Pr$\to$Rw & Rw$\to$Ar & Rw$\to$Cl & Rw$\to$Pr & Avg  \\
\midrule

Source train (RN50)     & \checkmark & 45.1 & 67.5 & 74.7 & 52.4 & 61.7 & 65.7 & 52.6 & 39.7 & 71.8 & 64.4 & 44.5 & 77.4 & 59.8 \\
G-SFDA (RN50) \cite{DBLP:journals/corr/abs-2108-01614} & \checkmark & 57.9  & 78.6  & 81.0  & 66.7  & 77.2 &  77.2  & 65.6  & 56.0 &  82.2 &  72.0  & 57.8  & 83.4  & 71.3 \\
CPGA (RN50) \cite{DBLP:journals/corr/abs-2106-15326} & \checkmark & 59.3 & 78.1 & 79.8 & 65.4 & 75.5 & 76.4 & 65.7 & 58.0 & 81.0 & 72.0 & 64.4 & 83.3 & 71.6 \\
SHOT (RN50)~\cite{liang2020we} & \checkmark & 57.1 & 78.1 & 81.5 & 68.0 & 78.2 & 78.1 & 67.4 & 54.9 & 82.2 & 73.3 & 58.8 & 84.3 & 71.8 \\
SHOT++ (RN50)~\cite{Spp} & \checkmark & 57.9 & 79.7 & 82.5 & 68.5 & 79.6 & 79.3 & 68.5 & 57.0 & 83.0 & 73.7 & 60.7 & 84.9 & 73.0 \\
\textbf{CoNMix} (RN50)           & \checkmark& 57.6 & 77.2 & 82.2 & 68.4 & 78.8 & 78.3 & 67.1 & 54.7 & 81.5 & 74.0 & 60.2 & 85.3 & 72.1\\
\midrule
\rowcolor{Gray}
Source train (DeiT-S)   & \checkmark & 51.7 & 74.2 & 79.3 & 62.6 & 72.5 & 74.7 & 64.0 & 47.5 & 79.6 & 69.9 & 49.8 & 80.9 & 67.2 \\
\rowcolor{Gray}
CDTrans (DeiT-S)~\cite{xu2021cdtrans} & \xmark & 60.6 & 79.5 & 82.4 & 75.6 & 81.0 & 82.3 & 72.5 & 56.7 & 84.4 & 77.0 & 59.1 & 85.5 & 74.7 \\
\rowcolor{Gray}
SHOT*(DeiT-S)~\cite{liang2020we} & \checkmark & 60.6 & 82.6 & 83.2 & 74.2 & 83.2 & 81.4 & 71.8 & 59.2 & 83.3 & 74.9 & 60.6 & 86.1 & 75.1 \\
\rowcolor{Gray}
SHOT++*(DeiT-S)~\cite{Spp} & \checkmark & 62.6 & 83.4 & 83.9 &	\textcolor{magenta}{\textbf{74.7}} & 83.3 &	\textcolor{magenta}{\textbf{82.7}} &	72.2 &	59.0 &	83.7 &	74.7 &	60.6 &	\textcolor{magenta}{\textbf{86.7}} &	75.6 \\
\rowcolor{Gray}
\textbf{CoNMix} (DeiT-S)         & \checkmark& \textcolor{magenta}{\textbf{63.4}} & \textcolor{magenta}{\textbf{83.5}} & \textcolor{magenta}{\textbf{84.6}} &                   73.7 & \textcolor{magenta}{\textbf{83.3}} & 82.2 & \textcolor{magenta}{\textbf{73.4}} & \textcolor{magenta}{\textbf{59.9}} & \textcolor{magenta}{\textbf{84.4}} & \textcolor{magenta}{\textbf{75.6}} & \textcolor{magenta}{\textbf{62.3}} & 85.9 & \textcolor{magenta}{\textbf{76.0}}\\
\bottomrule
 \end{tabular}}
\caption{\small Accuracy (\%) on Office-Home for SF-STDA. Methods that uses DeiT-S are compared within shaded region. * represents experiments implemented by us. \textit{CoNMix} (DeiT-S) achieves highest STDA average accuracy among all source-free methods.}
\label{tab:office_home_stda}
\end{table*}
\subsubsection{Source-free Multi-target Domain Adaptation}
\label{subsec:mtda}
For extending \textit{CoNMix} for SF-MTDA task, we propose a simple yet effective knowledge distillation based approach to transfer knowledge from all SF-STDA trained models (teachers) into a single student network (Stage 3 of Fig. \ref{fig:sfmtda_architecture}). Seminal work in KD by Hinton \etal ~\cite{hinton2015distilling} showed that the high temperature distillation is equivalent to minimizing $\mathcal{L}_{KD} = 0.5 (Z_{t} - Z_{l})^2$ loss which pays significant attention in matching the logits from two networks. However, simply using Hinton Loss tends to overfit the teacher predictions. To avoid memorization and sensitivity to training examples, we propose MixUp Knowledge Distillation inspired from Zhang \etal work \cite{zhang2017mixup}.
We first initialize a student model $g_{l}(x;\theta_l)$ with ImageNet trained weights. We store target image and it's corresponding pseudo label generated by each teacher network. An intermediate virtual domain image is generated by taking the convex combination of two randomly sampled images $\widetilde{x}_{ij} = \lambda x_i + (1-\lambda) x_j$ and $\widetilde{y}_{ij} = \lambda y_i + (1-\lambda) y_j$. Here $(x_i,y_i)$ represents image and pseudo label pairs sampled from $i^{th}$ domain. Here $\lambda \in [0,1]$. $(\widetilde{x}_{ij},\widetilde{y}_{ij})$ represents a sample from an intermediate domain. We use all such pairs to train the student network using as a knowledge distillation loss as shown in Eq. \ref{eq:l_mkd}. Derivation for Eq. \ref{eq:l_mkd} is shown in supplementary (see Section 2.4).
\small
\begin{align}\label{eq:l_mkd}
    \mathcal{L}_{MKD} &= \mathcal{L}_{CE}^{P_l}(\widetilde{x}_{ij},\widetilde{y}_{ij}) \nonumber\\
    \mathcal{L}_{MKD} &= \lambda \times \mathcal{L}_{CE}^{P_l}(\widetilde{x}_{ij}, y_i) + (1 - \lambda) \times \mathcal{L}_{CE}^{P_l}(\widetilde{x}_{ij}, y_j)
\end{align}
\normalsize
Intermediate domain acts as an implicit regularizer which helps to avoid over-fitting and generalize well on unlabeled target domains (Refer split domain test in supp. material). The proposed offline knowledge distillation allows us to distil knowledge from the best available STDA model because we aren't training teachers and students simultaneously. We can make inference with the final student model without needing domain labels. Refer suppl. material for more direct analysis on how multi domain features align.

\begin{table}[!htpb]
\scriptsize
\centering
\setlength{\tabcolsep}{3pt}
\resizebox{0.9\linewidth}{!}{%
\begin{tabular}{l|c|ccccccc}
\toprule
Method & SF &A$\to$D & A$\to$W & D$\to$A & D$\to$W & W$\to$A & W$\to$D & Avg\\
\midrule
Source train (RN50)     & \checkmark & 79.3 & 75.8 & 63.8 & 95.5 & 63.8 & 99.0 & 79.5\\ 
MA(RN50)~\cite{Li_2020_CVPR} & \checkmark & 92.7 &  93.7 &  75.3 & 98.5 & 77.8 & 99.8 & 89.6 \\
CPGA(RN50)~\cite{DBLP:journals/corr/abs-2106-15326} & \checkmark & 94.4 & \textcolor{magenta}{\textbf{94.1}} & 76.0 & 98.4 & 76.6 & 99.8 & 89.9 \\
SHOT(RN50)~\cite{liang2020we} & \checkmark & 94.0 & 90.1 & 74.7 &  98.4 & 74.3 & 99.9 & 88.6\\
SHOT(RN50)++~\cite{Spp} & \checkmark & \textcolor{magenta}{\textbf{94.3 }}& 90.4 & 76.2 & \textcolor{magenta}{\textbf{98.7}} & 75.8 & 99.9 & 89.2 \\
CoNMix (RN50)           & \checkmark & 88.8 & 94.0 & \textcolor{magenta}{\textbf{77.3}} & 98.1 & 75.2 & \textcolor{magenta}{\textbf{100.0}} & 88.9\\
\midrule
\rowcolor{Gray}
Source train (DeiT-S)   & \checkmark & 79.9 & 82.3 & 70.3 & 96.6 & 71.2 & 99.8 & 83.3\\
\rowcolor{Gray}
\rowcolor{Gray}
CDTrans (Deit-S)~\cite{xu2021cdtrans}& \xmark & 94.6 & 93.5 & 78.4 & 98.2 & \textcolor{magenta}{\textbf{78.0}} & 99.6 & \textcolor{magenta}{\textbf{90.4}} \\
\rowcolor{Gray}
CoNMix (DeiT-S)         & \checkmark & 90.6 & \textcolor{magenta}{\textbf{94.1}} & 77.2 & 98.1 & 77.0 & 99.6  &  89.4\\

\bottomrule
\end{tabular}}
\caption{\small Accuracy (\%) on Office-31 for STDA. Methods within shaded regions use DeiT backbone. }
\label{tab:office_31_stda}
\end{table}

\section{Experiments}
We conducted experiments using four popular benchmarking datasets: Office-31 \cite{saenko2010adapting}, Office-Home \cite{venkateswara2017deep} and large-scale like DomainNet \cite{peng2019moment} and VisDA \cite{peng2017visda} dataset.
After analyzing the benefits of VT over ResNet, we extended our analysis using VT as a backbone for \textit{CoNMix}. For a fair comparison, we conducted experiments on Office-31 and Office-Home using a smaller VT network (DeiT-S \cite{touvron2021training}) with 22M parameter for both student's and teacher's backbone because DeiT-S is comparable to ResNet50 (25M parameter). For DomainNet, we used Hybrid ViT \cite{dosovitskiy2020image} for teacher models in Stage-1 and Stage-2. In Stage-3, ResNet101 is used as student model. Please refer to suppl. for additional training details.
\label{sub:results_analysis}

\subsection{Evaluation}

\begin{table}[!htbp]
    \centering
    \setlength{\tabcolsep}{3pt}
    \resizebox{0.8\linewidth}{!}{
    \begin{tabular}{l|c|ccccc}
    \toprule
    & \multicolumn{1}{c|}{} && \multicolumn{3}{c}{Office-Home}\\
    \cline{3-7}
    Method & SF & Ar & Cl & Pr & Rw & Avg\\
    \midrule
    \rowcolor{Gray}
    Source only (RN50) & \checkmark & 62.5 & 61.2 & 55.1 & 61.8 & 60.1\\
    \rowcolor{Gray}
    Source only (DeiT-S) & \checkmark & 68.4 & 71.2 & 63.6 & 66.4 & 67.4\\
    \rowcolor{Gray}
    Domain-Aggregation & \checkmark &  69.5 &	77.2 &	66.4 &	67.0 &	70.0 \\
    \rowcolor{Gray}
    SHOT STDA in Stage-3 & \checkmark & 73.1	& 77.7	& 69.2	& 72.4	& 73.1 \\
    \rowcolor{Gray}
    CoNMix~(ours) & \checkmark & \textcolor{magenta}{\textbf{75.6}} & \textcolor{magenta}{\textbf{81.4}} & \textcolor{magenta}{\textbf{71.4}} & \textcolor{magenta}{\textbf{73.4}} & \textcolor{magenta}{\textbf{75.4}}\\
    \bottomrule
    \end{tabular}}
    \caption{\small SF-MTDA baselines. In \textit{Domain-Aggregation}, we combines multiple target domains and treat it as a single domain. In \textit{SHOT STDA}, we initialization for student network using SF-STDA SHOT weight. Highest performance for CoNMix highlights the importance of each design component in SF-MTDA.}
    \centering
    \label{tab:baselines}
\end{table}

\noindent\textbf{Results for SF-STDA:} We use Stage-2 of our proposed framework \textit{CoNMix} for the SF-STDA. Table.\ref{tab:office_home_stda} and \ref{tab:office_31_stda} illustrates results obtained for SF-STDA task for all combinations of domain pairs in Office-Home and Office-31. Our method outperforms existing source-free SOTA results with DeiT backbone in the case of the Office-Home dataset by a margin of \textbf{0.4\%}. We have achieved significant improvement for STDA in DomainNet by \textbf{6.0\%} (Refer supp. material table 1). Existing works \cite{yang2020heterogeneous,roy2021curriculum} only provide results for $Real$ and $Painting$ without comparing $Quickdraw$. We have compared against these works in Table.\ref{tab:DomainNet_Compare_acc}. However, we also report STDA accuracy for all other possible splits in suppl. material. Even though \textit{CoNMix} is source-free, we outperform non source-free method \cite{xu2021cdtrans} by \textbf{0.3\%} which showcases the efficacy of the proposed approach. We have included SF-STDA results for VisDA datasets in suppl. material (Table 4).

\begin{table}[!b]
    \scriptsize
    \centering
    \def\arraystretch{1.1}
    \setlength{\tabcolsep}{1.5 pt}
    \resizebox{\linewidth}{!}{\begin{tabular}{l|c|ccccccccc}
    \specialrule{1pt}{1pt}{1pt}
    & & \multicolumn{9}{c}{DomainNet} \\
    \cline{3-11}
    Model & SF & R $\rightarrow$ S & R $\rightarrow$ C & R $\rightarrow$ I & R $\rightarrow$ P & P $\rightarrow$ S & P $\rightarrow$ R & P $\rightarrow$ C & P $\rightarrow$ I &  \textbf{Avg} (\%)\\
    \specialrule{1pt}{1pt}{1pt}
    CDAN~\cite{long2018conditional}         & \xmark & 40.7 & 51.9 & 22.5 & 49.0 & 39.6 & 57.9 & 44.6 & 18.4 & 40.6 \\
    HGAN~\cite{yang2020heterogeneous}       & \xmark & 34.3 & 43.2 & 17.8 & 43.4 & 35.7 & 52.3 & 35.9 & 15.6 & 34.7 \\
    {CDAN} + {DCL}~\cite{roy2021curriculum} & \xmark & 45.2 & 58.0 & 23.7 & 54.0 & 45.0 & 61.5 & 50.7 & 20.3 & 44.8 \\
    {\domours}~\cite{roy2021curriculum}     & \xmark & 48.4 & 59.6 & 25.3 & 55.6 & 45.3 & 58.2 & 51.0 & 21.7 & 45.6 \\
    
    \midrule
    \rowcolor{Gray}
    CoNMix~(ours) & \checkmark & \textcolor{magenta}{\textbf{52.9}} & \textcolor{magenta}{\textbf{63.5}} & \textcolor{magenta}{\textbf{27.7}} & \textcolor{magenta}{\textbf{59.5}} & \textcolor{magenta}{\textbf{53.3}} & \textcolor{magenta}{\textbf{71.8}} & \textcolor{magenta}{\textbf{59.7}} & \textcolor{magenta}{\textbf{24.0}} & \textcolor{magenta}{\textbf{51.6}} \\  
    \specialrule{1pt}{1pt}{1pt}
    \end{tabular}}
    \caption{\small \% Accuracy for SF-STDA on DomainNet Dataset.  Our source-free method (Shaded region) outperforms the existing SOTA with significant margin even though they access source-dataset during target adaptation.}
    \label{tab:DomainNet_Compare_acc}
\end{table}
\noindent\textbf{Results for SF-MTDA:} There are no existing comprehensive studies related to SF-MTDA. Therefore, we formulated few baselines to evaluate SF-MTDA and reported the results in the Table. \ref{tab:baselines}. We have considered source only training as a initial baseline, where we train on only on source dataset and evaluate its performance on all the target domains.  For source train row in Table. \ref{tab:baselines} Art ($Ar$) represents training on $Ar$ domain and testing on remaining domains. From Table. \ref{tab:baselines}, we can observe that we achieve test accuracy of 60.1 \% and 67.4\% using ResNet and DeiT-S backbone respectively. Since, these results do not incorporate any adaptation, therefore, performing any adaptation using these models should lead to improvement in accuracy. Secondly, we considered aggregating all the target domain datasets together and train SF-STDA model. We can see its performance in Row-3 \textit{(Domain-Aggregation)} of Table. \ref{tab:baselines} is better than source only but lesser than CoNMix. It shows that the proposed training strategy for CoNMix utilises domain information effectively. In another baseline (Row-4: \textit{SHOT STDA in Stage-3}, we perform student training using SHOT SF-STDA weights in place of CoNMix SF-STDA weights. We can observe that its performance lies between domain aggregation and CoNMix. Therefore, the main components in SF-MTDA such as proposed teacher training and MixUp plays an important role in achieving the desired result.
Our SF-MTDA results on popular benchmark datasets will serve as a new baseline for research in this direction. Each cell in Table \ref{tab:SOTA-Office31-Home} and \ref{tab:MTDA_DomainNet} reports classification accuracy of model which is adapted from \textit{Source Domain} $\rightarrow$ \textit{Rest of Target Domains}. SF represents whether the algorithm supports source-free method or not.
We fine-tune the student network using MKD objective on all the target domains.
Experiment with ResNet-101 provides initial baseline which consists of source training using ResNet-101 backbone and directly evaluating its performance on target dataset without performing any adaptation.
Our source-free method achieves a significant improvement of \textbf{5.2\%} over existing SOTA methods on the Office-Home dataset even though other methods access the labeled source data during adaptation. Our experiments with DomainNet dataset can be used to validate the scalability of our SF-MTDA (Table \ref{tab:MTDA_DomainNet}). We are the first to provide results for large-scale DomainNet dataset for both SF-STDA and SF-MTDA. 

\begin{table}[!htbp]
    \label{tab:MTDA}
    \centering
    \small
    \def\arraystretch{}
    \setlength{\tabcolsep}{3pt}
    \resizebox{\linewidth}{!}{\begin{tabular}{l|c|cccccccccc}
    \specialrule{1.5pt}{1pt}{1pt}
    \multicolumn{1}{c|}{} && \multicolumn{5}{c}{Office-31} && \multicolumn{4}{c}{Office-Home}\\
    \cline{3-6} \cline{8-12}
    Model & SF & A & D & W & \textbf{Avg}(\%) & & Ar & Cl & Pr & Rw & \textbf{Avg}(\%) \\
    \specialrule{1.5pt}{1pt}{1pt}
    RevGrad~\cite{ganin2016domain} & \xmark & 78.2 & 72.2 & 69.8 & 73.4 & & 58.4 & 58.1 & 52.9 & 62.1 & 57.9\\
    CDAN~\cite{long2018conditional} & \xmark & 93.6	& 80.5	& 81.3	& 85.1 & & 59.5 & 61.0 & 54.7	& 62.9& 59.5\\
    AMEAN~\cite{chen2019blending} & \xmark & 90.1 & 77.0 & 73.4 & 80.2 & & 64.3 & 65.5 & 59.5 & 66.7 & 64.0\\
    MT-MTDA~\cite{nguyen2021unsupervised} & \xmark & 87.9 & 83.7 & 84.0 & 85.2 & & 64.6 & 66.4 & 59.2 & 67.1 & 64.3\\
    HGAN~\cite{yang2020heterogeneous} & \xmark & 88.0 & 84.4 & 84.9 & 85.8 & & - & - & - & - & -\\
    {CGCT}~\cite{roy2021curriculum}    & \xmark & \textcolor{magenta}{\textbf{93.9}} & 85.1 & 85.6 & 88.2 & & 67.4 & 68.1 & 61.6 & 68.7 & 66.5\\
    {D-CGCT}~\cite{roy2021curriculum}  & \xmark & 93.4 & \textcolor{magenta}{\textbf{86.0 }}& \textcolor{magenta}{\textbf{87.1 }}& \textcolor{magenta}{\textbf{88.8 }}& & 70.5 & 71.6 & 66.0 & 71.2 & 69.8\\
    \midrule
    \rowcolor{Gray}
    Source train (RN50) & \checkmark & 76.3 & 68.7 & 67.0 & 70.7 & & 62.5 & 61.2 & 55.1 & 61.8 & 60.1\\
    \rowcolor{Gray}
    Source train (DeiT-S) & \checkmark & 81.4 & 76.1 & 75.5 & 77.7 & & 68.4 & 71.2 & 63.6 & 66.4 & 67.4\\
    \rowcolor{Gray}
     CoNMix~(ours) & \checkmark & 92.4 & 81.8 & 80.4 & 84.9 & & \textcolor{magenta}{\textbf{75.6}} & \textcolor{magenta}{\textbf{81.4}} & \textcolor{magenta}{\textbf{71.4}} & \textcolor{magenta}{\textbf{73.4}} & \textcolor{magenta}{\textbf{75.4}}\\

    \specialrule{1.5pt}{1pt}{1pt}
    \end{tabular}}
    \caption{\small \% Accuracy for Office-31 and Office-Home dataset for SF-MTDA. \textit{CoNMix} outperforms SOTA in all possible splits of Office-Home.}
    \label{tab:SOTA-Office31-Home}
\end{table}

\begin{table}[!ht]
\begin{minipage}{0.45\linewidth}
\setlength{\tabcolsep}{3pt}
\resizebox{\linewidth}{!}{
\begin{tabular}{l|c|ccccc}
\toprule
& \multicolumn{1}{c|}{} && \multicolumn{3}{c}{Office-Caltech}\\
\cline{3-7}
Method & SF & A & C & D & W & Avg\\
\midrule
ResNet-101 \cite{he2016deep}        & \xmark & 90.5 & 94.3 & 88.7 & 82.5 & 89.0 \\
SE \cite{french2018self}            & \xmark & 90.3 & 94.7 & 88.5 & 85.3 & 89.7 \\
MCD \cite{saito2018maximum}         & \xmark & 91.7 & 95.3 & 89.5 & 84.3 & 90.2 \\
DANN \cite{ganin2015unsupervised}   & \xmark & 91.5 & 94.3 & 90.5 & 86.3 & 90.7 \\
DADA \cite{peng2019domain}          & \xmark & 92.0 & 95.1 & 91.3 & 93.1 & 92.9 \\
\midrule
\rowcolor{Gray}
Source only                   & \checkmark & 90.7 & 96.1 & 90.2 & 90.9 & 92.0 \\
\rowcolor{Gray}
SHOT                    & \checkmark & 96.2 & 97.3 & 96.3 & 96.2 & 96.5 \\
\rowcolor{Gray}
\textbf{CoNMix}~(ours)  & \checkmark & \textbf{\color{magenta}96.4} & \textbf{\color{magenta}97.4} & \textbf{\color{magenta}96.9} & \textbf{\color{magenta}96.8} & \textbf{\color{magenta}96.9} \\
\bottomrule
\end{tabular}
}
\end{minipage}
\quad
\begin{minipage}{0.5\linewidth}

\setlength{\tabcolsep}{3pt}
\resizebox{\linewidth}{!}{%
\begin{tabular}{l|c|ccccccc}
\toprule
& \multicolumn{1}{c|}{} && \multicolumn{3}{c}{DomainNet}\\
\cline{3-9}
Method & SF & Cli.	& Inf.	& Pai. & Qui. & Rea. & Ske. & Avg\\
\midrule
SE~\cite{french2018self} & \xmark & 21.3&8.5&14.5&13.8&16.0&19.7&15.6 \\
MCD~\cite{saito2018maximum} & \xmark & 25.1&19.1&27.0&10.4&20.2&22.5&20.7 \\
DADA~\cite{peng2019domain} & \xmark & 26.1&20.0& 26.5&12.9&20.7&22.8&21.5 \\
CDAN~\cite{long2018conditional} & \xmark & 31.6&27.1&31.8&12.5&33.2&35.8&28.7 \\
MCC~\cite{jin2020minimum} & \xmark & 33.6&30.0&32.4&13.5&28.0& 35.3& 28.8\\
CGCT~\cite{roy2021curriculum} & \xmark  & 36.1& {33.3}&35.0&10.0&39.6&39.7&32.3\\
D-CGCT~\cite{roy2021curriculum} & \xmark & {37.0}&32.2&{37.3}&19.3&{39.8}&40.8&{34.4}\\
\rowcolor{Gray}
\midrule
Source (RN101) & \checkmark & 25.6 & 16.8&25.8&9.2&20.6&22.3&20.1 \\
\rowcolor{Gray}
\textbf{CoNMix}~(ours) & \checkmark & \textcolor{magenta}{\textbf{41.8}} & \textcolor{magenta}{\textbf{29.2}} & \textcolor{magenta}{\textbf{39.9}} & \textcolor{magenta}{\textbf{17.5}} & \textcolor{magenta}{\textbf{32.7}} & \textcolor{magenta}{\textbf{41.2}} & \textcolor{magenta}{\textbf{33.7}}\\
\bottomrule
\end{tabular}
}
\end{minipage}
\caption{\small Classification accuracy (\%) on Office-Caltech and DomainNet for MTDA. Methods in shaded region are source-free.}
\label{tab:MTDA_DomainNet}
\end{table}
\noindent\textbf{Ablation for loss function:}
For better understanding of the effect of each loss function, we conducted an ablation study for different losses and show the result in  Table \ref{tab:ablation}. If we use $\mathcal{L}_{CE}^{P_l}$ or $\mathcal{L}_{Cons}$ individually, the performance is very poor. We observe that the $\mathcal{L}_{NM}$ is the most important loss component in overall optimization objective. Relative gains due to $\mathcal{L}_{CE}^{P_l}$ and $\mathcal{L}_{cons}$ may be smaller but we achieve best performance when all the components are present. The usage of both  $\mathcal{L}_{CE}^{P_l}$ and $\mathcal{L}_{cons}$ together is not expected to handle noise present in pseudo label during training and it will deteriorate the model performance. These observations are consistent across various splits. We add additional analysis on pseudo label refinement and loss function in supplementary material (supp. Fig. 1 \& 5).

\begin{table}[!htp]
\centering
\setlength{\tabcolsep}{3pt}
\resizebox{\linewidth}{!}{\begin{tabular}{l|l|l|cccccc}
\toprule
$\mathcal{L}_{NM}$ & $\mathcal{L}_{Cons}$ & $\mathcal{L}_{CE}^{P_l}$ & Ar$\rightarrow$Cl & Cl$\rightarrow$Ar & Pr$\rightarrow$Cl & Cl$\rightarrow$Pr & Rw$\rightarrow$Cl & Cl$\rightarrow$Rw \\
\midrule
 &  & \checkmark & 6.3 & 15.7 & 8.8 & 5.5 & 2.9 & 13.9 \\ 
 & \checkmark &  & 3.8 & 6.7 & 4.1 & 1.2 & 4.3 & 5.4 \\ 
\checkmark &  &  & 59.5 & 71.1 & 55.9 & 78.6 & 58.3 & 78.7 \\ 
\checkmark & \checkmark &  & 60.3 & 72.7 & 57.6 & 80.4 &59.8  & 78.6 \\
\checkmark & \checkmark & \checkmark & 63.8 & 73.7 & 59.9 & 83.3 & 62.3 & 82.2 \\
\bottomrule
\end{tabular}}
\caption{\small Ablation for target accuracy when various losses are introduced sequentially across various splits of Office-Home.}
\label{tab:ablation}
\end{table}

\section{Conclusion}
In this work, we introduced a novel framework (\textit{CoNMix}) for solving source-free Single and Multi-target domain adaptation. We achieved SOTA results for various datasets and in some cases CoNMix performed better than even non-source free methods. We provided baseline for source-free STDA and MTDA methods on DomainNet for the first time, which can help the domain adaptation research community further investigate this novel direction. Further, we provided empirical insights along with quantitative and qualitative results highlighting the benefit of VT and suggest that VT could be a potential choice for feature extractor in designing novel domain adaptation algorithms. We showed that our design choice, such as Nuclear-Norm Maximization, consistency constraint and label refinement mitigate uncertainty associated with noisy labels. \textit{CoNMix} demonstrated effectiveness through various experimental findings across datasets, therefore we are keen to extend this further for more challenging source-free online adaptation where target domains are dynamic and continuously changing.

\noindent \textbf{Acknowledgement}: This work is supported by a Young Scientist Research Award (Sanction no. 59/20/11/2020-BRNS) from DAE-BRNS, India.

{\small
\bibliographystyle{ieee_fullname}
\bibliography{egbib}
}

\end{document}